\documentclass[
]{ceurart}

\usepackage{listings}
\begin{document}

\copyrightyear{2025}
\copyrightclause{Copyright for this paper by its authors.
  Use permitted under Creative Commons License Attribution 4.0
  International (CC BY 4.0).}

\conference{IberLEF 2025, September 2025, Zaragoza, Spain}

\title{CardiffNLP at CLEARS-2025: Prompting Large Language Models for Plain Language and Easy-to-Read Text Rewriting}

\author[1]{Mutaz Ayesh}[%
email=AyeshMA@cardiff.ac.uk
]
\address[1]{Cardiff University, Wales, UK}
\author[1]{Nicolás Gutiérrez-Rolón}[%
email=GutierrezRolonN@cardiff.ac.uk
]
\author[1]{Fernando Alva-Manchego}[%
email=AlvaManchegoF@cardiff.ac.uk
]

\begin{abstract}
  This paper details the CardiffNLP team's contribution to the CLEARS shared task on Spanish text adaptation, hosted by IberLEF 2025. The shared task contained two subtasks and the team submitted to both. Our team took an LLM-prompting approach with different prompt variations. While we initially experimented with LLaMA-3.2, we adopted Gemma-3 for our final submission, and landed third place in Subtask 1 and second place in Subtask 2. We detail our numerous prompt variations, examples, and experimental results.
\end{abstract}


\maketitle

\section{Introduction}

Access to clear and understandable information constitutes an important facet of the right to information, as determined in the Universal Declaration for Human Rights \citep{udhr}. It enables individuals to make informed decisions, participate in a democracy, and exercise other fundamental rights, yet many public and official documents are still inaccessible to significant parts of the population due to their technical language, especially in the public sector and fields like law and medicine. The CLEARS shared task \citep{clears_at_iberlef2025} at IberLEF-2025 \citep{iberlef2025overview} is intended to address this challenge; it tackles automatic adaptation of Spanish language texts into two accessible formats: Plain Language (PL) and Easy-to-Read (E2R). 

These two subtasks aim to support different target populations and special needs. PL targets general audiences regardless of reading level or background, and can help both non-native speakers and people with reading limitations. It promotes clarity and conciseness by using active voice, as well as common and everyday words to eliminate ambiguity, and avoiding jargon. E2R, on the other hand, is designed according to the UNE 153101 EX guidelines \citep{UNE_2018} to meet the needs of people with cognitive, intellectual, or learning disabilities, such as dyslexia, ADHD, and aphasia, among others. It also emphasizes structural and linguistic simplicity by reducing cognitive load, using short and simple sentences and clear and direct language, and generally involves the target audience in testing and reviewing texts to ensure clarity and ease of understanding \citep{martínez2024exploringlargelanguagemodels}. E2R-adapted texts can also take the form of short sentences with simple vocabulary and grammar, including explaining complex concepts in simpler terms and splitting sentences into separate lines \citep{calleja-etal-2024-automating}.

Traditionally, the creation of these types of texts is a manual and very resource-intensive process, involving experts of several disciplines, as well as validators and users to check the readability of the texts. As the volume of information grows, it becomes impractical to rely solely on human experts to create these adaptations, sparking the interest of scientists and researchers in the field of Natural Language Processing (NLP) to develop automatic systems capable of producing these simplifications with minimal human intervention.

In our submission, we explored a range of prompting-based methods that leverage the power of Large Language Models (LLMs), specifically LLaMA 3 \citep{llama3modelcard}, and Gemma 3 \citep{gemma_2025}, to perform automatic text adaptation. We experimented with zero-shot, one-shot, and few-shot prompting strategies to assess the effectiveness of these methods and the quality of their results for both settings of PL and E2R adaptations. Our results highlight the potential of these models, as well as limitations, and complexity of evaluating text outputs with a variety of metrics.

\section{Related Work}

%Multiple papers that addressed automatic simplification of Spanish texts incorporated a qualitative approach to evaluate their texts. 
\citet{martínez2024exploringlargelanguagemodels} investigated the use of LLaMA-2 models (specifically the 7B version with and without QLoRA fine-tuning, and the 70B version) for simplification of Spanish texts in line with E2R guidelines. Their experiments used the \textit{Amas Fácil Foundation} corpus, which consists of 1,941 sentence pairs of original and E2R-adapted texts. One of their approaches involved translating the original text to English, then simplifying that version with LLaMA, and then translating the output back into Spanish. Qualitative evaluation was carried out with the help of professional E2R adapters and individuals with intellectual disabilities. They found that the 70B version of LLaMA-2 with translation yielded the best results, outperforming the 7B version fine-tuned on the sentence pairs. We build on this work by experimenting with LLaMA models and applying the same E2R guidelines in the prompts.

\citet{towardsreliabletexts} proposed a threefold evaluation framework for E2R texts: automatic metrics to measure text complexity quantitatively, a checklist-based manual evaluation addressing sentence-level and format-level aspects along with meaning preservation, and finally a user-centered evaluation focused on the needs of users and the understandability of texts. Their study emphasized gathering direct feedback from readers by asking participants to retell the stories they read and respond to questions about the simplified texts. We incorporate aspects of their approach by integrating sentence-level, format-level and meaning-preservation guidelines into our prompts. 

\citet{calleja-etal-2024-automating} examined both quantitative (F1; Sigma; and BLEU) and qualitative (human) evaluations for E2R text segmentation. Their study compared multiple methodologies: first, scoring-based segmentation via constituency parsing (using Universal Dependencies) and masked language model scoring (using mBERT), as well as prompting-based segmentation with LLaMA-2-7B (zero- and few-shot) and QLoRA finetuning. Among these methodologies, unsupervised MLM-based segmentation proved most effective for Spanish. We take inspiration from their prompt variations.

Finally, \citet{Ponce_2024} investigated generating meaning-preserving sentence sequences from a single complex input sentence using zero- and few-shot prompting and fine-tuning with datasets of varying complexity. Their results showed that LLM models fine-tuned with LoRA achieved the strongest performance overall, although these struggled to consistently follow their splitting instructions. They also evaluated their results using quantitative and qualitative evaluation, where the conclusions drawn from automated metric results were supported by two human evaluations. Inspired by this, our prompts explicitly guide the model to consider sentence-based splits in a ``reading stage'' prior to simplification.

% [add a concluding sentence]
\section{Dataset Description}

The CLEARS corpus \citep{espinosa-zaragoza-etal-2023-automatic,botella2024clearsim} consists of 3,000 pieces of news from various municipalities in the province of Alicante in Spain, covering a variety of topics including sports, culture, leisure, and festivities. For each article, a PL version and a E2R version were generated. The dataset was divided into a 70/30 training/testing split, consisting of 2,100 news articles for training and 900 news articles for testing. 

Our experiments used different subsets of the training data as examples in the prompt in the one- and few-shot settings.

\section{System Overview and Experimental Setup}

Our experiments were mainly conducted on the PL dataset. We started with Meta's 3B parameter instruction-tuned version of Llama 3.2, then later adopted the 4B parameter instruction-tuned version of Google's Gemma 3 in our final submission.\footnote{We also briefly experimented with EuroLLM, though it produced suboptimal texts, often with non-Spanish and ungrammatical words.} The experiments included zero-, one-, and few-shot prompting, and numerous variants of the user and system messages passed to the model, which also included experimenting with the language in which these messages were written, whether in English or Spanish, to determine its effect on performance.
%The simplification of each variant was obtained using a Python function that streamlines the process by taking in the complex sentence and multiple other arguments (depending on the prompting type) and yields the generated text.

\subsection{Metrics}
Given the resource-exhausting nature of prompting using LLMs, the evaluation of each variant of the prompts was conducted on a subset of 100 sentences from the provided PL dataset. We computed: (1) the SentenceBERT cosine similarity between the generated simplification and its complex sentence and reference simplification, separately, and (2) the Fernández-Huerta score of the generated text. 

The mean cosine similarity score of the complex and simple sentences and the mean Fernández-Huerta score of the reference simplifications were used as baselines and served as scores against which the performance of each variant was compared. To ensure robustness, we extended our evaluation to a randomly selected set of 200 sentences to ensure that the performance of the variant will not worsen when applied on the rest of the dataset. Additional metrics included BERTScore (F1), as well as the evaluation script provided by the organizers.

These metrics do not fully capture the nuances of text simplification, particularly for E2R. A significant aspect of E2R is the visual formatting and sentence segmentation for readability. None of the given automatic metrics have any way of measuring or addressing this. This highlights a persistent challenge in the field of evaluating text simplification tasks.

\subsection{Preliminary Experiments}
We followed an iterative process in which the generated output was compared to the reference simplification for analysis, and then the prompt was modified accordingly. In each iteration, new instructions were added to the prompt to address errors in format, factuality, grammaticality, and compliance with instructions. We observed that instructing the model to return its output as a Python dictionary yielded the best results, and facilitated the extraction of the simplification using a Python function. Furthermore, explicitly instructing the model to read the sentence first and to work on it on a sentence-level helped increase similarity scores with respect to both complex and simple references.

Most prompt engineering was initially conducted using LLaMA-3.2. However, once performance plateaued and the same mistakes kept arising (e.g. the model returning the input complex sentence as its generated simplification, returning an English simplification, returning one of the examples as a simplification, or returning an answer not formatted as a Python dictionary), we switched to Gemma-3. 
Gemma-3 consistently performed as well or better than LLaMA-3.2, showing greater compliance with the formatting instructions. The model also showed superior performance when prompted in Spanish, as English prompts led the model to simplify the complex sentence in English.

\subsubsection{Zero-shot prompting}
Zero-shot prompting was attempted once, mainly due to unreliable and unpredictable output formats. LLaMA-3.2 would split long paragraphs into shorter chunks of paragraphs with headlines, like the one in Appendix \ref{prompt1zerosample1}, or a short paragraph like in \ref{prompt1zerosample2}. In addition, the factuality of the generated simplification was questioned, with LLM ``hallucinations'' being the most prominent in dates and paraphrased numbers (``2000 people'' being changed to ``many people''). The generated output also included the use of first-person pronouns, as if the model was narrating the text. Later prompts included explicitly instructing the model not to make any of these mistakes in the messages. The prompts and sample generated simplifications can be found in Appendix \ref{appendix:zeroshot}.

\subsubsection{One-shot prompting}

Since the format of the generated text outputted by zero-shot prompting was unreliable and diverse, providing the model with an example of what an ideal simplification should look like was essential. One-shot prompting was carried out in three separate prompts, dubbed P2, P3, and P4. Notably, P4 is distinguished by the integration of the E2R guidelines used in \citep{martínez2024exploringlargelanguagemodels} into the prompt. Those guidelines were extracted from ``the Spanish standard and European Guidelines in Spanish ILSMH (1998)''. The Spanish versions of the messages and guidelines were manually translated by us. Messages and sample simplifications can be found in Appendix \ref{appendix:oneshot}, while the guidelines  can be found in Appendix \ref{appendix:guidelines}.

\subsubsection{Few-shot prompting}

In this setting, three examples were included in the user role message. All few-shot prompts included the guidelines whether in English or Spanish. Preliminary results showed that prompts and guidelines written in English obtained better results in LLaMA-3.2, and therefore English was adopted for LLaMA-3.2. When Gemma-3 was used, the messages and guidelines written in Spanish showed better results (compare P3-4 with P5-7 in Table \ref{tab:gemma}) and were therefore used in our experiments and final submission. Messages and sample simplifications can be found in Appendix \ref{appendix:fewshot}. Another improvement to the few-shot LLaMA-3.2 prompts (P5-7, Table \ref{tab:llamafewshot}) and Gemma-3 prompts (P5-7, Table \ref{tab:gemma}) was the instruction of returning a Python dictionary rather than free-form text,% which  The improvement in the results can be seen in those tables.

The same mistakes detailed in one-shot prompting persisted in few-shot prompting, especially the ones pertaining to the model returning the examples provided as its generated simplification. This issue persisted even when the model was explicitly instructed not to do so and told to only learn from the format and style of the simplifications in the examples, as in Prompt 6 (Appendix \ref{appendix:fewshot}). Despite the incremental change in the scores, this issue drove us to consider Gemma-3. 

Particularly, despite the prompt reinforcing not to include phrases like ``Here is your simplification'', the model was inconsistent and sometimes provided these preambles. The use of a Python dictionary format mitigated many of these errors; however, formatting inconsistencies still occurred. Some of these included the use of colons instead of equal signs for key-value assignment, as well as irregular spacing, and minor deviations in terminology that required additional post-processing. Although these issues did not fundamentally affect the simplification, it produced noise that was picked up by the metrics, and necessitated custom cleaning functions to address and standardise the output.

\subsection{Final Submission}

The prompt selected for submission was P7 using Gemma-3, which achieved the best performance in most metrics. The system message was slightly modified depending on the subtask, primarily by adapting the few-shot examples to match the characteristics of the reference simplification as present in the training data. Both variants retained the instruction to return  a Python dictionary and incorporated the refined guidelines. The final prompt ensured great structural consistency, reduced hallucinations, and improved alignment with the reference simplification, which can be seen in Table \ref{tab:test}. 

\section{Results}

The preliminary results described previously can be found in Tables \ref{tab:llamazerooneshot}, \ref{tab:llamafewshot}, \ref{tab:gemma}, and \ref{tab:e2rfinal}, while the final results when P7 was applied on the test set can be found in Table \ref{tab:test}.

% [To add final results @Nico]

\textbf{Subtask 1 - PL}. The organizers averaged cosine similarity scores obtained using TF-IDF embeddings and using BERT embeddings. In the former we achieved 63\% and in the latter 77\%, yielding a 70\% cosine similarity score, landing us at number 3. This makes sense as we did not take TF-IDF into account in our evaluation. The highest average similarity score was 75\% achieved by HULAT-UC3M. A notable feat in this regard is our cosine similarity score obtained by BERT embeddings which was slightly outperformed by HULAT-UC3M's 78\%. In terms of the Fernández-Huerta complexity metric, we landed at number 2 with a complexity score of 78.81, only outperformed by VICOMTECH with their 82.98. Our score indicates a "somewhat easy" readability level that matches that of sixth graders.

\textbf{Subtask 2 - E2R}. The same approach was used for this task, with our TF-IDF-based (65\%) and BERT-based (77\%) cosine similarity scores being averaged to yield the final 71\% score. This score landed us at second place, only slightly outperformed by NIL-UCM's 72\%. In terms of the Fernández-Huerta complexity metric, however, we landed in third place, with an averaged 77.85 complexity score.

\begin{table}[]
    \centering
    \caption{Preliminary scores for zero- (P1) and one-shot (P2-4) Llama-3-based models on a 100-sentence subset from the PL dataset. "CS" refers to SentenceBERT cosine similarity.}
    \begin{tabular}{|c|c|c|ccc|}
         \hline
         \textbf{Metric} & \textbf{Reference simple} & \textbf{P1} & \textbf{P2} & \textbf{P3} & \textbf{P4}\\
         \hline
         \textbf{Mean CS with complex} & 
         0.8459 & 0.6897 & 0.6833 & 0.8018 & 0.7382\\
         \textbf{Mean CS with simple} & 
         N/A & 0.6889 & 0.6775 & 0.8143 & 0.7218\\
         \hline
         \textbf{Mean Fernández-Huerta score} &
         80.84 & 93.19 & 81.46 & 87.81 & 93.51\\
         \hline
    \end{tabular}
    
    \label{tab:llamazerooneshot}
\end{table}

\begin{table}[]
    \centering
    \caption{Preliminary scores for few-shot Llama-3-based models on a 200-sentence subset from the PL dataset. "CS" refers to SentenceBERT cosine similarity, while "CLEARS sim." refers to the cosine similarity function provided evaluation script by the organizers.}
    \begin{tabular}{|c|c|ccc|}
         \hline
         \textbf{Metric} & \textbf{Reference simple} & \textbf{P5} & \textbf{P6} & \textbf{P7}\\
         \hline
         \textbf{Mean CS with complex} & 0.846 & 0.8222 & 0.8365 & 0.8738\\
         \textbf{Mean CS with simple} & N/A & 0.8211 & 0.8223 & 0.8232\\
         \hline
         \textbf{Mean Fernández-Huerta score} & 80.831 & 87.65 & 83.71 & 75.99\\
         \hline
         \textbf{Mean CLEARS sim. with complex} & 0.860 & 0.8243 & 0.8479 & 0.8794\\
         \textbf{Mean CLEARS sim. with simple} & N/A & 0.7984 & 0.8067 & 0.8093\\
         \hline
         \textbf{Mean BERTScore (F1) with complex} & 0.753 & 0.7469 & 0.7638 & 0.7897\\
         \textbf{Mean BERTScore (F1) with simple} & N/A & 0.7360 & 0.7364 & 0.7294\\
         \hline
    \end{tabular}
    
    \label{tab:llamafewshot}
\end{table}

\begin{table}[]
    \centering
    \caption{Preliminary scores for Gemma-3-based models on a 200-sentence subset from the PL dataset. All prompts used few-shot prompting. P3-4 used English prompts, P4 included the guidelines. P5-7 used Spanish prompts and guidelines.}
    \begin{tabular}{|c|c|cc|ccc|}
         \hline
         \textbf{Metric} & \textbf{Ref. simple} & \textbf{P3} & \textbf{P4} & \textbf{P5} & \textbf{P6} & \textbf{P7}\\
         \hline
         \textbf{Mean SentenceBERT cosine with complex} &
         0.8459 & 0.7910 & 0.7870 & 0.8298 & 0.8540 & 0.8468\\
         \textbf{Mean SentenceBERT cosine with simple} & 
         N/A & 0.8003 & 0.8053 & 0.8390 & 0.8558 & 0.8552\\
         \hline
         \textbf{Mean BERTScore (F1) with complex} &
         0.7532 & 0.7272 & 0.7402 & 0.7561 & 0.7695 & 0.7669\\
         \textbf{Mean BERTScore (F1) with simple} &
         N/A & 0.7535 & 0.7616 & 0.7742 & 0.7837 & 0.7758\\
         \hline
    \end{tabular}
    
    \label{tab:gemma}
\end{table}

\begin{table}[]
    \centering
    \caption{Preliminary scores for Gemma-3-based models on a sample of 100 sentences from the E2R training dataset.}
    \begin{tabular}{|c|c|c|}
         \hline
         \textbf{Metric} & \textbf{Ref. simple} & \textbf{P7} \\
         \hline
         \textbf{Mean SentenceBERT cosine with complex} &
         0.8331 & 0.8387\\
         \textbf{Mean SentenceBERT cosine with simple} & 
         N/A & 0.85189\\
         \hline
         \textbf{Mean Fernández-Huerta score} & 79.73 & 90.06\\
         \hline
         \textbf{Mean BERTScore (f1) with complex} & 0.7476 & 0.7613\\
         \textbf{Mean BERTScore (F1) with simple} &
         N/A & 0.7722\\
         \hline
    \end{tabular}
    
    \label{tab:e2rfinal}
\end{table}

\begin{table}[]
    \centering
    \caption{Final scores of P7 applied on the test set. The sample was a subset of 100 sentences, while the others columns reflect the results of the entire test set.}
    \begin{tabular}{|c|c|cc|}
    \hline
    \textbf{Metric} & \textbf{Sample} & \textbf{Subtask 1} & \textbf{Subtask 2}\\
    \hline
    Mean CS with reference & 0.8276 & 0.8199 & 0.8319\\
    \hline
    Mean Fernandez-Huerta & 91.03 & 91.66 & 91.61\\
    \hline
    Mean BERT Score & 0.7546 & 0.7589 & 0.7724\\
    \hline
    \end{tabular}
    
    \label{tab:test}
\end{table}

\section{Conclusion}

For our submission, we explored a prompting-based approach to automatic text simplification using instruction-tuned LLMs for the CLEARS shared task, focusing on Plain Language and Easy-to-Read adaptations in Spanish. Through an iterative experimental process, we evaluated zero-, one-, and few-shot prompting strategies with LLaMA-3.2 and Gemma-3. Our results show that, while both models demonstrated considerable potential in text simplification, Gemma-3 outperformed in terms of adherence to formatting and language accuracy, particularly when prompting directly in Spanish. 

Adding sentence-level and explicit linguistic instructions, in addition to enforcing a structured output format (i.e., Python dictionary format) proved crucial in mitigating common LLM errors, such as hallucinations and inconsistent formatting. These strategies allowed for more reliable extraction and evaluation of simplifications and contributed to improved alignment with reference texts.

Overall, contributing to the CLEARS shared task was a productive endeavor that deepened our understanding of the capabilities and limitations of instruction-tuned LLMs in the context of text simplification, as well as the nuances of evaluating automatic generated outputs from these models. While the current prompting-based methods show promising results, the work also highlights ongoing challenges in the field, particularly for E2R outputs where formatting and readability go beyond what current automatic metrics can adequately capture. The results also highlight the importance of a human in the loop, as the instructions provided to the models were based on careful and repetitive evaluation of the generated output. Future work will benefit from incorporating human evaluation and developing metrics that better reflect the complex qualitative aspects of simplification, especially one that requires intricate formatting for specific target groups.

%% Define the bibliography file to be used
\bibliography{bibliography}

\begin{thebibliography}{12}
\expandafter\ifx\csname natexlab\endcsname\relax\def\natexlab#1{#1}\fi
\providecommand{\url}[1]{\texttt{#1}}
\providecommand{\href}[2]{#2}
\providecommand{\path}[1]{#1}
\providecommand{\DOIprefix}{doi:}
\providecommand{\ArXivprefix}{arXiv:}
\providecommand{\URLprefix}{URL: }
\providecommand{\Pubmedprefix}{pmid:}
\providecommand{\doi}[1]{\href{http://dx.doi.org/#1}{\path{#1}}}
\providecommand{\Pubmed}[1]{\href{pmid:#1}{\path{#1}}}
\providecommand{\bibinfo}[2]{#2}
\ifx\xfnm\relax \def\xfnm[#1]{\unskip,\space#1}\fi
%Type = Misc
\bibitem[{{United Nations}(1948)}]{udhr}
\bibinfo{author}{{United Nations}}, \bibinfo{title}{Universal declaration of human rights}, \bibinfo{year}{1948}.
%Type = Article
\bibitem[{Botella-Gil et~al.(2025)Botella-Gil, Espinosa-Zaragoza, Bonet-Jover, Madina, Molino~Pi{\~n}ar, Moreda, Gonzalez-Dios, Mart{\'i}n~Valdivia, and Ure{\~n}a}]{clears_at_iberlef2025}
\bibinfo{author}{B.~Botella-Gil}, \bibinfo{author}{I.~Espinosa-Zaragoza}, \bibinfo{author}{A.~Bonet-Jover}, \bibinfo{author}{M.~Madina}, \bibinfo{author}{L.~Molino~Pi{\~n}ar}, \bibinfo{author}{P.~Moreda}, \bibinfo{author}{I.~Gonzalez-Dios}, \bibinfo{author}{M.~T. Mart{\'i}n~Valdivia}, \bibinfo{author}{Ure{\~n}a},
\newblock \bibinfo{title}{Overview of clears at iberlef 2025: Challenge for plain language and easy-to-read adaptation for spanish texts},
\newblock \bibinfo{journal}{Procesamiento del Lenguaje Natural} \bibinfo{volume}{75} (\bibinfo{year}{2025}).
%Type = Inproceedings
\bibitem[{Gonz{\'a}lez-Barba et~al.(2025)Gonz{\'a}lez-Barba, Chiruzzo, and Jim{\'e}nez-Zafra}]{iberlef2025overview}
\bibinfo{author}{J.~{\'A}. Gonz{\'a}lez-Barba}, \bibinfo{author}{L.~Chiruzzo}, \bibinfo{author}{S.~M. Jim{\'e}nez-Zafra},
\newblock \bibinfo{title}{{Overview of IberLEF 2025: Natural Language Processing Challenges for Spanish and other Iberian Languages}},
\newblock in: \bibinfo{booktitle}{Proceedings of the Iberian Languages Evaluation Forum (IberLEF 2025), co-located with the 41st Conference of the Spanish Society for Natural Language Processing (SEPLN 2025), CEUR-WS. org}, \bibinfo{year}{2025}.
%Type = Misc
\bibitem[{{UNE}(2018)}]{UNE_2018}
\bibinfo{author}{{UNE}}, \bibinfo{title}{Une 153101:2018: Easy-to-read standards}, \bibinfo{year}{2018}.
%Type = Misc
\bibitem[{Martínez et~al.(2024)Martínez, Moreno, and Ramos}]{martínez2024exploringlargelanguagemodels}
\bibinfo{author}{P.~Martínez}, \bibinfo{author}{L.~Moreno}, \bibinfo{author}{A.~Ramos}, \bibinfo{title}{Exploring large language models to generate easy to read content}, \bibinfo{year}{2024}. \URLprefix \url{https://arxiv.org/abs/2407.20046}. \href{http://arxiv.org/abs/2407.20046}{{\tt arXiv:2407.20046}}.
%Type = Inproceedings
\bibitem[{Calleja et~al.(2024)Calleja, Etchegoyhen, and Ponce}]{calleja-etal-2024-automating}
\bibinfo{author}{J.~Calleja}, \bibinfo{author}{T.~Etchegoyhen}, \bibinfo{author}{D.~Ponce},
\newblock \bibinfo{title}{Automating easy read text segmentation},
\newblock in: \bibinfo{editor}{Y.~Al-Onaizan}, \bibinfo{editor}{M.~Bansal}, \bibinfo{editor}{Y.-N. Chen} (Eds.), \bibinfo{booktitle}{Findings of the Association for Computational Linguistics: EMNLP 2024}, \bibinfo{publisher}{Association for Computational Linguistics}, \bibinfo{address}{Miami, Florida, USA}, \bibinfo{year}{2024}, pp. \bibinfo{pages}{11876--11894}. \URLprefix \url{https://aclanthology.org/2024.findings-emnlp.694/}. \DOIprefix\doi{10.18653/v1/2024.findings-emnlp.694}.
%Type = Article
\bibitem[{AI@Meta(2024)}]{llama3modelcard}
\bibinfo{author}{AI@Meta},
\newblock \bibinfo{title}{Llama 3 model card}  (\bibinfo{year}{2024}). \URLprefix \url{https://github.com/meta-llama/llama3/blob/main/MODEL_CARD.md}.
%Type = Article
\bibitem[{Team(2025)}]{gemma_2025}
\bibinfo{author}{G.~Team},
\newblock \bibinfo{title}{Gemma 3}  (\bibinfo{year}{2025}). \URLprefix \url{https://goo.gle/Gemma3Report}.
%Type = Inproceedings
\bibitem[{Madina et~al.(2024)Madina, Gonzalez-Dios, and Siegel}]{towardsreliabletexts}
\bibinfo{author}{M.~Madina}, \bibinfo{author}{I.~Gonzalez-Dios}, \bibinfo{author}{M.~Siegel},
\newblock \bibinfo{title}{Towards reliable e2r texts: A proposal for standardized evaluation practices},
\newblock in: \bibinfo{editor}{K.~Miesenberger}, \bibinfo{editor}{P.~Pe{\v{n}}{\'a}z}, \bibinfo{editor}{M.~Kobayashi} (Eds.), \bibinfo{booktitle}{Computers Helping People with Special Needs}, \bibinfo{publisher}{Springer Nature Switzerland}, \bibinfo{address}{Cham}, \bibinfo{year}{2024}, pp. \bibinfo{pages}{224--231}.
%Type = Inproceedings
\bibitem[{Ponce et~al.(2024)Ponce, Etchegoyhen, Calleja, and Gete}]{Ponce_2024}
\bibinfo{author}{D.~Ponce}, \bibinfo{author}{T.~Etchegoyhen}, \bibinfo{author}{J.~Calleja}, \bibinfo{author}{H.~Gete},
\newblock \bibinfo{title}{Split and rephrase with large language models},
\newblock in: \bibinfo{booktitle}{Proceedings of the 62nd Annual Meeting of the Association for Computational Linguistics (Volume 1: Long Papers)}, \bibinfo{publisher}{Association for Computational Linguistics}, \bibinfo{year}{2024}, p. \bibinfo{pages}{11588–11607}. \URLprefix \url{http://dx.doi.org/10.18653/v1/2024.acl-long.622}. \DOIprefix\doi{10.18653/v1/2024.acl-long.622}.
%Type = Inproceedings
\bibitem[{Espinosa-Zaragoza et~al.(2023)Espinosa-Zaragoza, Abreu-Salas, Moreda, and Palomar}]{espinosa-zaragoza-etal-2023-automatic}
\bibinfo{author}{I.~Espinosa-Zaragoza}, \bibinfo{author}{J.~Abreu-Salas}, \bibinfo{author}{P.~Moreda}, \bibinfo{author}{M.~Palomar},
\newblock \bibinfo{title}{Automatic text simplification for people with cognitive disabilities: Resource creation within the {C}lear{T}ext project},
\newblock in: \bibinfo{editor}{S.~{\v{S}}tajner}, \bibinfo{editor}{H.~Saggio}, \bibinfo{editor}{M.~Shardlow}, \bibinfo{editor}{F.~Alva-Manchego} (Eds.), \bibinfo{booktitle}{Proceedings of the Second Workshop on Text Simplification, Accessibility and Readability}, \bibinfo{publisher}{INCOMA Ltd., Shoumen, Bulgaria}, \bibinfo{address}{Varna, Bulgaria}, \bibinfo{year}{2023}, pp. \bibinfo{pages}{68--77}. \URLprefix \url{https://aclanthology.org/2023.tsar-1.7/}.
%Type = Misc
\bibitem[{Botella-Gil et~al.(2024)Botella-Gil, Espinosa-Zaragoza, Moreda, and Palomar}]{botella2024clearsim}
\bibinfo{author}{B.~Botella-Gil}, \bibinfo{author}{I.~Espinosa-Zaragoza}, \bibinfo{author}{P.~Moreda}, \bibinfo{author}{M.~Palomar}, \bibinfo{title}{{Corpus ClearSim}}, \bibinfo{year}{2024}. \URLprefix \url{http://hdl.handle.net/10045/151688}.

\end{thebibliography}

\appendix

\section{Guidelines}

\textit{Guidelines in English}: 

"Here are some guidelines: Linked ideas should be separated by a period instead of a comma. The semicolon (;) should not be used. Use simple and commonly used language. Avoid using abstract, technical, or complex terms. Avoid using words in other languages unless they are widely known and properly explained. Deter from using expressions or metaphors that all readers may not understand, unless they are common in everyday language. Use the same word throughout the text to refer to the same object or referent. Use simple sentences and avoid complex sentence. Use the present indicative whenever possible. Avoid compound or uncommon verb tenses, as well as the use of conditionals and subjunctives. Avoid using the passive voice. Avoid the use of impersonal sentences. Avoid using two or more verbs in a row, except for periphrases with modal verbs like 'should', 'want', 'know', and 'can'. Use affirmative sentences, except in cases such as simple prohibitions, where negative forms may be clearer and more direct. Avoid negative forms and double negations. Include only one main idea in each sentence."

\textit{Guidelines in Spanish}: 

"Aquí hay unas pautas. Las ideas vinculadas deben separarse con un punto en lugar de una coma.
No se debe usar el punto y coma (;). Se debe usar lenguaje simple y de uso común. El vocabulario debe ser adecuado para el usuario del documento.
Se debe evitar usar términos abstractos, técnicos o complejos, al igual que palabras en otros idiomas a menos que sean ampliamente conocidas y estén adecuadamente explicadas.
Se debe evitar usar metáforas o expresiones que no todos los lectores puedan entender a menos que sean comunes en el lenguaje cotidiano.
Se debe usar la mismo palabra en todo el texto para hacer referencia al mismo objeto o referente.
Se debe usar oraciones simples y evitar oraciones complejas.
Se debe usar el presente indicativo siempre que sea posible.
Se deben evitar tiempos verbales complejos, compuestos o poco comunes, tal y como el uso de condicionales y subjuntivos.
Se debe evitar el uso de la voz pasiva.
Se debe evitar el uso de oraciones impersonales.
Se debe evitar usar dos o más verbos seguidos, excepto para perífrasis con verbos modales como 'deber', 'querer', 'saber' y 'poder'.
Se debe usar oraciones afirmativas, excepto en casos como prohibiciones simples, donde las formas negativas pueden ser más claras y más directas.
Se debe evitar las formas negativas y las dobles negaciones. Incluye solo una idea principal en cada oración."

\label{appendix:guidelines}

\section{PL — LLaMA-3.2 prompts}

\subsection{Zero-shot prompting (Prompt 1)}

Prompt 1 was a zero-shot prompt. These prompts were also translated into Spanish.

\textit{System message}: You are a professional language editor and simplifier. You transform sentences to make them easier to understand for people with intellectual disabilities and difficulties in reading comprehension. You use very simple, short, direct sentences in the active voice, and avoid complicated words. You do not add information that does not appear in the original text."

\textit{User role message}: "Here is your complicated sentence. Transform it to make it easier to understand for people with intellectual disabilities and difficulties in reading comprehension. Use very simple, short, direct sentences in the active voice, and avoid complicated words. Do not add information that does not appear in the original sentence. When you identify a date such as a the date when an event is taking place, keep it as is. Do not use first-person pronouns in your simplification unless they are present in the text. Do not paraphrase numbers; meaning, do not change "2000 people" to "many people". Here is an example of a complex sentence: \_\_\_\_ and its simplification: \_\_\_\_. Your output should resemble that format. Here is your complicated sentence: \_\_\_\_."

\label{appendix:zeroshot}

\subsubsection{Sample 1}

\textit{Complex sentence}: 

"Alicante, 10 de noviembre del 2022. Alicante inicia la cuenta atrás con los preparativos que van a clausurar los actos conmemorativos del Centenario del Mercado Central. El Ayuntamiento engalana con un montaje especial de jardines verticales y columnas florales el emblemático edificio, que el próximo sábado cumple sus cien años con un evento en el que se va a proyectar un videomapping y se celebra un concierto a partir de las 19 horas.Los alumnos del ciclo formativo de jardinería y floristería del IES Virgen del Remedio, el arquitecto Jose Botí, Tina Floristas y la Asociación de Comerciantes del Mercado Central, han sido los que han realizado durante las últimas semanas la creación del diseño y elaboración de las estructuras para crear la decoración floral colgante que se va a instalar en la cúpula y los pórticos para el día del acto del Centenario, cuyo montaje ya se ha iniciado y se puede ver en la fachada de la plaza 25 de Mayo.La concejala de Mercados, Lidia López, ha destacado “el gran resultado y diseño florar que se ha elaborado y que pone en valor el edificio del Mercado Central”, al tiempo que agradeció a “todas las personas que de forma altruista han trabajado intensamente en el diseño de la decoración floral para que la fachada de nuestro Mercado Central luzca festiva el día de su Centenario”.Esta semana se han ultimado los trabajos de la decoración ornamental vegetal del Mercado Central, en la que la parte principal del diseño es el montaje en la cúpula de un jardín vertical con plantas artificiales colgantes que caerán desde la misma y además con cien bombillas que simularán una lámpara haciendo referencia a los cien años del Mercado. Montaje que se instalará con una grúa especial y trabajaos de altura para ponerlos en la cúpula. También se está trabajando en la decoración de la fachada del emblemático edificio, en la que lucirán arcos y columnas florales en los pórticos de la facha de la plaza 25 de mayo realizados con plantas naturales.El presidente de la Asociación de Mercados, Paco Alemañ, puso en valor “la implicación, dedicación y esfuerzo de todos los alumnos del IES Virgen del Remedio y las personas que han colaborado desde el primer momento como el arquitecto Jose Botí y Tina Floristas para sacar adelanta la decoración del Mercado. Estoy seguro que la decoración floral va a lucir mucho y gustar a todos los que vengan a celebrar el Centenario, así como agradezco la implicación del Ayuntamiento de Alicante que se ha volcado en todos los actos y actividades organizados estos meses”.La edil de Mercado ha recordado que “esta semana ya estamos con el cierre de todos los actos del Centenario del Mercado Central, está siendo unos días muy intensos de actividades y como acto final se celebrará una fiesta el sábado con la proyección de un videomapping y el concierto de Varry Brava, a cuál invitamos a todos los ciudadanos que quieran acompañarnos que tendrá lugar delante de la fachada principal en la avenida Alfonso X el Sábio”Acompañando a la decoración floral este sábado el Mercado Central clausura las actividades de su Centenario con dos conciertos de entrada libre y la proyección de un videomapping en la fachada del Mercado Central a partir de las 19 horas. Varry Brava es el grupo que pondrá el broche final al Centenario en la avenida Alfonso X El Sabio, y como complemento actuará el grupo ‘Uña y Carne’."

\textit{Reference simplification}:

"Alicante se prepara para terminar los actos de celebración del Centenario del Mercado Central con un evento especial el próximo sábado. El Ayuntamiento ha decorado el edificio con jardines verticales y columnas de flores. Los alumnos del Instituto de Educación Secundaria Virgen del Remedio, el arquitecto José Botí, Tina Floristas y la Asociación de Comerciantes del Mercado Central crearon la decoración floral colgante que se instalará en la cúpula y las puertas para el día del acto del Centenario. La concejala de Mercados, Lidia López, ha elogiado el resultado y diseño floral. La decoración con adornos incluye un jardín vertical con plantas artificiales colgantes y 100 bombillas que simularán una lámpara en referencia a los 100 años del Mercado. También habrá arcos y columnas de flores en la fachada del edificio. El presidente de la Asociación de Mercados agradeció la participación de todas las personas que colaboraron en la decoración. El evento del sábado contará con la proyección de un videomapping a las 7 de la tarde y a continuación 2 conciertos gratuitos de Varry Brava y el grupo Uña y Carne.
Palabras difíciles
Videomapping es proyectar imágenes o videos
sobre paredes, edificios y otras superficies.
"

\textit{Generated simplification}: 

"\textbf{Alicante celebra el Centenario del Mercado Central}

El Ayuntamiento de Alicante ha empezado a decorar el Mercado Central con flores y plantas para celebrar su centenario. Los estudiantes del IES Virgen del Remedio y otros grupos han trabajado duro para hacerlo.

\textbf{La decoración}

En la cúpula del Mercado Central, se ha montado un jardín vertical con plantas artificiales que caerán de la cúpula. También se han colocado cien bombillas que simulan una lámpara. En la fachada del Mercado, se han decorado los pórticos con arcos y columnas florales.

\textbf{Conciertos y videomapping}

El próximo sábado, el Mercado Central cerrará con dos conciertos de entrada libre y la proyección de un videomapping en la fachada del Mercado Central. El grupo Varry Brava actuará en la avenida Alfonso X El Sabio, acompañado del grupo 'Uña y Carne'.

\textbf{Agradecimientos}

La concejala de Mercados, Lidia López, ha agradecido a todos los que han trabajado en la decoración floral. También ha destacado el gran resultado del diseño floral que se ha elaborado.

\textbf{La celebración}

El presidente de la Asociación de Mercados, Paco Alemañ, ha expresado su gratitud a todos los que han colaborado en la decoración del Mercado. Está seguro de que la decoración floral será hermosa y agradecerá a todos los que vengan a celebrar el centenario."

\label{prompt1zerosample1}

\subsubsection{Sample 2}

\textit{Complex sentence}: La Concejalía de Cultura del Ayuntamiento de Benidorm ha programado para el próximos sábado 6 y el viernes 12 dos actividades dirigidas a público de todas las edades que se celebrarán en el Parque de Foietes.La primera de ellas tendrá lugar el sábado 6 a las 21 horas y se trata del espectáculo musical infantil y familiar "Tararos" a cargo de la compañía Trébol Teatro. La entrada es gratuita.La segunda será el próximo 12 de agosto, viernes, y tendrá como protagonista al Mag Malastruc con su show de magia. En este caso el inicio será también a las 21 horas y la entrada será libre para todos.

\textit{Reference simplification}: La Concejalía de Cultura de Benidorm ha organizado un espectÃ¡culo musical infantil el sábado 6 de agosto a las 9 de la noche llamado Tararos y un espectáculo de magia con Mag Malastruc para el viernes 12 de agosto a las 9 de la noche. Los dos eventos son gratis. 

\textit{Generated simplification}: "La Concejalía de Cultura de Benidorm ha organizado un espectáculo musical infantil el sábado 6 de agosto a las 9 de la noche llamado Tararos y un espectáculo de magia con Mag Malastruc para el viernes 12 de agosto a las 9 de la noche. Los dos eventos son gratis."

\label{prompt1zerosample2}

\subsection{One-shot prompting}

\subsubsection{Prompt 2}

\textit{System message}: "You are a professional language editor and simplifier. You transform sentences to make them easier to understand for people with intellectual disabilities and difficulties in reading comprehension. You use very simple, short, direct sentences in the active voice, and avoid complicated words. You do not add information that does not appear in the original text."

\textit{User role message}: "Transform the complex sentence to make it easier to understand for people with intellectual disabilities and difficulties in reading comprehension. Use very simple, short, direct sentences in the active voice, and avoid complicated words. Do not add information that does not appear in the original sentence. When you identify a date such as a the date when an event is taking place, keep it as is. Do not use first-person pronouns in your simplification unless they are present in the text. Do not paraphrase numbers; meaning, do not change "2000 people" to "many people". Here is an example of a complex sentence: \_\_\_\_ and its simplification: \_\_\_\_. Your output should resemble that format. Here is your complicated sentence: \_\_\_\_"

\subsubsection{Prompt 3}

\textit{System message}: "You are a professional language editor and simplifier. You transform sentences to make them easier to understand for people with intellectual disabilities and difficulties in reading comprehension. You use very simple, short, direct sentences in the active voice, and avoid complicated words. You are honest about the facts: you do not add information that does not appear in the original text, and you include proper names and dates in your simplification without altering or removing them."

\textit{User role message}: "Here is your sentence. Transform it to make it easier to understand for people with intellectual disabilities and difficulties in reading comprehension. Use very simple, short, direct sentences in the active voice, and avoid complicated words. Do not add information that does not appear in the original sentence. When you identify a date such as a the date when an event is taking place, keep it as is. Do not use first-person pronouns in your simplification unless they are present in the text. Do not paraphrase numbers; meaning, do not change "2000" people to "many people". Years and ranges of years are important and should be kept when detected. For the sake of producing a factual simplification, do not skip proper nouns like place names or personal names. Here is an example of a complex sentence: \_\_\_\_ and its simplification: \_\_\_\_. Here is your new complicated sentence to simplify: \_\_\_\_."

\subsubsection{Prompt 4 - adding the guidelines to the prompt}

\textit{System message}: "You are a professional language editor and simplifier. You transform sentences to make them easier to understand for people with intellectual disabilities and difficulties in reading comprehension. You use very simple, short, direct sentences in the active voice, and avoid complicated words. [English guidelines from Appendix \ref{appendix:guidelines}]."

\textit{User role message}: "Here is your sentence. Transform it to make it easier to understand for people with intellectual disabilities and difficulties in reading comprehension. Use very simple, short, direct sentences in the active voice, and avoid complicated words. Do not add information that does not appear in the original sentence. \textit{[Guidelines]}.  For the sake of producing a factual simplification, do not skip proper nouns like place names or personal names. When you identify a date such as a the date when an event is taking place, keep it as is. Do not use first-person pronouns in your simplification unless they are present in the text. Do not paraphrase numbers; meaning, do not change "2000" people to "many people". Years and ranges of years are important and should be kept when detected. Here is an example of a complex sentence: \_\_\_\_ and its simplification: \_\_\_\_. Your output should resemble that format. Here is your new complicated sentence: \_\_\_\_."

\textit{Sample complex sentence}: "Alicante, 23 de noviembre de 2022.- El Museo de Arte Contemporáneo de Alicante, vuelve a celebrar, en su sexta edición Todos MACA, un encuentro dirigido a todos los públicos y todas las edades, cuya entrada es gratuita y totalmente accesible.El evento cuenta con la colaboración de la Concejalía de Cultura que dirige Antonio Manresa que ha destacado que "se trata de unas jornadas artísticas multidisciplinares con el objetivo de visibilizar y empoderar a las personas con discapacidad. Y lo hace con los mismos objetivos de siempre porque, por desgracia, sigue siendo necesario trabajar en ello: que todas las personas puedan participar de los procesos habituales creativos, de los circuitos artísticos, y estar presentes en todos los centros educativos, en todos los espacios profesionales y en cada estrato de esta sociedad".Según señalan desde el museo "Buscamos la equidistancia social, política y humana. Algo que no ha llegado todavía por muchas cuestiones. Y mientras se mantenga ese “todavía”, aquí estaremos, un año más desde 2017, celebrando estos encuentros e intercambios colmados de emoción e inclusión en el MACA en torno al Día Internacional de las Personas con Discapacidad que se celebra cada 3 de diciembre"."Con Todos MACA, continua Manresa, lo que buscamos es acercar el arte contemporáneo a todo tipo de público, propiciando así el interés por la cultura, las artes, la participación, la creatividad y el respeto de todos los colectivos y estratos de la ciudadanía. La idea es tender puentes entre las personas, y hacer de nuestras diferencias, de nuestra diversidad, en lugar de un obstáculo o un impedimento comunicativo, una oportunidad de aprendizaje mutuo, de creación colectiva. Todos sumamos si todos participamos".El programa de Todos MACA se compone de numerosas actividades artísticas y culturales distribuidas a lo largo de un fin de semana (2, 3 y 4 de diciembre), en el que convertiremos el museo en un espacio privilegiado para despertar el genio de un espectador ajeno al mundo del arte, estableciendo un diálogo con el público e insertando el museo en el entramado social de la ciudad para convertirse en un lugar para todos.El público destinatario de estas jornadas es toda la sociedad, sin el cual difícilmente se podría construir un mundo más inclusivo, tolerante y empático si no es contando con las personas que lo componen. Por eso, el programa de actividades no entiende de colectivos concretos: está abierto a todas las personas.El evento está coordinado por Extrarradio, Arte e Inclusión Social y DaGuten, Escuela Gráfica de Barrio. Dos colectivos de educadores, mediadores y artistas interdisciplinares que organizan desde hace casi una década toda clase de ideas, proyectos, sueños y locuras maravillosas en la que Todos MACA está incluida.ProgramaciónViernes 2 de diciembre11 horas. Presentación "Una colección en lectura fácil".Para dar comienzo a TODOS MACA 2022 tendrá lugar la presentación de un folleto, resultado del trabajo del Equipo Seri en el proyecto “Tirando del hilo”. Este proyecto educativo y artístico diseñado por DaGuten. Escuela Gráfica de Barrio está dirigido a artistas del Centro Ocupacional Terramar de APSA. Durante este año el Equipo Seri ha diseñado un folleto en lectura fácil para cada una de las colecciones del MACA, y queremos presentarlo en la Colección Arte Siglo XX. Y es que el proyecto Tirando del hilo ha sido algo más que una serie de talleres de artes gráficas. El Equipo Seri será el encargado de explicar la Colección y la técnica de la serigrafía para que cada uno/a de los participantes se puedan llevar a casa una obra inspirada en esta Colección.18 horas. Visita guiada a una visita sin vista.La Colección Arte Siglo XX es una de las mayores joyas de esta ciudad. Un lugar idóneo para aprender, descubrir, inspirarse y, sobre todo, respirar arte. Por eso, tendrá lugar una visita guiada muy especial que se realizará sin el sentido de la vista, para poder explorar el resto (oído, gusto, olfato, tacto) en una experiencia que busca la estimulación sensorial y la empatía con aquellas personas con discapacidad visual.Sábado 3 de diciembre12 horas. Visita guiada "Grandes artistas a secas". Con Intérprete de Lengua de Signos Española (LSE).Además, se celebrará la visita-taller con vermú guiada por Explicarte, pero esta vez con servicio de interpretación de lengua de signos española, para poder ampliar el público objetivo de la misma. Estará centrada en Frida Kahlo, artista atenta a las preocupaciones plásticas de su generación y dominada por su contexto sociocultural. De ella se conocerá, no sólo su obra, sino también su vida a través de la discapacidad, con la intención de despertar una nueva mirada, que reconozca el trabajo de la mujer en la superación de los roles y modelos impuestos en la historia del arte.18 horas. Proyección cortos de animaciónDurante 15 minutos se proyectarán en el salón de actos del MACA, 2 cortometrajes de animación sobre discapacidad, educación inclusiva y superación para disfrutar en familia. Tras la proyección, los visitantes se podrán llevar una serigrafía de “La tragedia del volcán”, seleccionado en 19º Festival de Cine de Alicante en la categoría de animación y de "Sempere Motion", que surge de la invitación al MACA por parte de CreArt a participar en el Día Europeo de la Creatividad Artística 2018.Domingo 4 de diciembre11.30 horas. Juegos/ taller seri y "Tiro porque nos toca a todos"El domingo concluirá con el juego colectivo “Juego de la Oca de Los Seres Diversos”. Una oportunidad para participar en este fantástico proyecto colaborativo que tiende puentes entre personas y para jugar a ser artistas. Tendrá lugar en la Plaza de Santa María, en frente del MACA, donde se desarrollará el juego.Toda la información sobre las inscripciones a las actividades en www.maca-alicante.es"

\textit{Reference simplification}: ``El Museo de Arte Contemporáneo de Alicante celebra la edición número 6 de Todos MACA, un evento gratuito y accesible para todos los públicos y todas las edades en colaboración con la Concejalía de Cultura. El objetivo del encuentro es hacer visible y dar más autonomía a las personas con discapacidad y aumentar la participación en los procesos creativos y artísticos. El programa de actividades del evento incluye numerosas actividades artísticas y culturales realizadas a lo largo de un fin de semana. Las actividades están dirigidas a toda la sociedad y buscan tender puentes entre las personas, haciendo de nuestras diferencias una oportunidad de aprendizaje de todos. El evento está coordinado por Extrarradio, Arte e Inclusión Social y DaGuten, Escuela Gráfica de Barrio. Viernes 2 de diciembre, 11 de la mañana. Presentación Una colección en lectura fácil. 6 de la tarde. Visita guiada a una visita sin vista.  Sábado 3 de diciembre, 12 de la mañana. Visita guiada Grandes artistas a seca. Con Intérprete de Lengua de Signos Española (LSE). 6 de la tarde. Proyección cortos de animación. Domingo 4 de diciembre, 11 y media de la mañana. Juegos/ taller seri yTiro porque nos toca a todos.''

\textit{Sample simplification generated by LLaMA-3.2}: ``El Museo de Arte Contemporáneo de Alicante está celebrando su sexta edición, Todos MACA. El evento es gratuito y accesible para todos. El Ayuntamiento de Alicante ha colaborado con el museo para hacer que el evento sea más inclusivo. El evento cuenta con actividades artísticas y culturales para todos los públicos. El programa incluye una visita guiada sin vista, una visita guiada con interpretación de lengua de signos española y una proyección de cortos de animación. También habrá juegos y talleres para todos los públicos. El evento es una oportunidad para acercar el arte contemporáneo a todos y promover la inclusión y la empatía.''

\label{appendix:oneshot}

\subsection{Few-shot prompting}

\subsubsection{Prompt 5}

No sentence-level instruction.

\textit{System message}: ``You are a professional language editor and simplifier. You transform sentences to make them easier to understand for people with intellectual disabilities and difficulties in reading comprehension. You do not add information that does not appear in the text. Your aim is two-fold: you want to simplify the text by identifying each sentence and then simplifying it, and to produce a simplified text that is factual and that retains the important parts. Important parts like proper nouns (like place names like cities or countries; personal names; event names) should appear in your simplification. You do not paraphrase numbers; meaning, do not change ``2000'' people to ``many people''. Years and ranges of years are important and should be kept. When you identify a date in any format, keep it as is. You include all relevant information. You do not use first-person pronouns such as "I" or "we" in your simplification, unless they are present in the text. [English guidelines from Appendix \ref{appendix:guidelines}].''

\textit{User role message}: ``You are now provided with three pairs of complex-simple sentences as examples; the simplified sentences were simplified by experts like yourself. Here is your first example of a complex sentence ``\_\_\_\_'' and its simplification: ``\_\_\_\_''. Here is your second example of a complex sentence: ``\_\_\_\_'' and its simplification: ``\_\_\_\_''. Here is your third example of a complex sentence: ``\_\_\_\_'' and its simplification: ``\_\_\_\_''. They are for you to learn from in terms of the style, length, and overall structure of the simplified sentences, not to use in your simplification! Here is your new complex sentence: ``\_\_\_\_''. Provide its simplification. Assume that your simplification will be stored as a value to a key called ``simple'' in a Python dictionary. Your output should be the dictionary value and nothing else.''

\subsubsection{Prompt 6}

Instructed the model to work on a sentence-level.

\textit{System message}: ``You are a professional language editor and simplifier. You transform sentences to make them easier to understand for people with intellectual disabilities and difficulties in reading comprehension. You work on a sentence-level; meaning, you detect sentences, simplify them, then concatenate all the simplified sentences together. You use very simple, short, direct sentences in the active voice, and avoid complicated words. You do not add information that does not appear in the text. Your aim is two-fold: you want to simplify the text by identifying each sentence and then simplifying it, and to produce a simplified text that is factual and that retains the important parts. Important parts like proper nouns (like place names like cities or countries; personal names; event names) should appear in your simplification. You do not paraphrase numbers; meaning, do not change ``2000'' people to "many people". Years and ranges of years are important and should be kept. When you identify a date in any format, keep it as is. You include all relevant information. You do not use first-person pronouns such as ``I'' or ``we'' in your simplification, unless they are present in the text. [English guidelines from Appendix \ref{appendix:guidelines}].''

\textit{User role message}: ``You are now provided with three pairs of complex-simple sentences as examples; the simplified sentences provided were simplified by experts like yourself. They are provided for you to learn from in terms of the style, length, and overall structure, not to use in your simplification! Here is your first example of a complex sentence "\_\_\_\_" and its simplification: "\_\_\_\_". Here is your second example of a complex sentence: "{}" and its simplification: "\_\_\_\_". Here is your third example of a complex sentence: "\_\_\_\_" and its simplification: "\_\_\_\_". Here is your new complex sentence: "\_\_\_\_". Provide its simplification. You using a sentence-level simplification, so you should retain the most amount of information as possible. Assume that your simplification will be stored as a value to a key called ``simple'' in a Python dictionary. Your output should be the dictionary value and nothing else.''

\textit{Sample complex sentence}: "La Societat Musical ‘La Nova’ interpretará este domingo, 4 de diciembre, su tradicional Concierto de Santa Cecilia en el Auditorio del Centro Cultural, a las 11.30h. Así lo ha anunciado hoy el concejal de Cultura, Jaime Jesús Pérez, junto al presidente de la entidad, Juan Vicente Moreno y  su director titular, Francisco José Rovira Peretó.Pérez ha valorado la colaboración especial de ‘La Nova’ que “llena de actividad cultural” la programación navideña de Benidorm, especialmente en 2022, año en el que, la más joven de las sociedades musicales de la ciudad ha cumplido sus primeros diez años.El concejal de Cultura ha dado la enhorabuena por las actividades organizadas con motivo del aniversario y ha recordado que su Escuela de Música “cuenta con más de 300 educandos” y que durante el concierto de homenaje a Santa Cecilia se incorporarán once nuevos músicos a la Banda Sinfónica que se sumarán a los 80 actuales. ‘La Nova’ cuenta además con la Banda de la Escuela de Música, formada por medio centenar de intérpretes y su masa social la conforman más de 300 socios.En su intervención, Rovira Peretó, ha recordado los eventos que han tenido lugar con motivo del décimo aniversario, como el concierto en el Auditorio de la Diputación de Alicante, o el Festival de Bandas de octubre en el que intervinieron diversas sociedades musicales hermanadas con ‘La Nova’. Sobre el concierto de Santa Cecilia ha señalado que contará con una primera parte a cargo de la Banda de la Escuela de Música y una segunda de la Banda Sinfónica.Entre las piezas que interpretará la banda, Francisco José Rovera Peretó destacó ‘El Quijote’, fantasía sinfónica de Ferrer Ferran con la narración de José Mª Albero, que fue “la primera actuación pública” de la 'La Nova' hace ahora diez años. El director también ha añadido que el día 10 tendrá lugar un concierto de Navidad en el que “se presentará la banda sonora de la Cabalgata de Reyes”, el CD ‘De Oriente a Benidorm’ y, el 17 de diciembre, interpretará otro concierto en la Parroquia de La Almudena.Por último, Juan Vicente Moreno ha animado a “todo el pueblo” a presenciar los conciertos y anunció, ya en febrero, la celebración de un concierto de marchas procesionales en la Iglesia del Buen Pastor, organizado en colaboración con la Cofradía de la Buena Esperanza."

\textit{Reference simplification}: "La Societat Musical La Nova dará el Concierto de Santa Cecilia el domingo 4 de diciembre a las 11 y media de la mañana en el Auditorio del Centro Cultural. Este año La Societat Musical La Nova participa en la programación de la Navidad de Benidorm porque cumple 10 años. En el concierto participarán los 80 músicos actuales de la banda y 11 músicos nuevos. Una de las obras musicales que tocarán en el concierto será El Quijote de Ferrer Ferran donde habla José María Albero. También habrá otros conciertos como uno de Navidad y otro en la Parroquia de la Almudena. El presidente de la Sociedad Musical La Nova Juan Vicente Moreno animó a todo el pueblo a ir a ver los conciertos."

\textit{Generated simplification}: "La Societat Musical 'La Nova' interpretará su Concierto de Santa Cecilia el 4 de diciembre. El concejal de Cultura ha valorado la colaboración de 'La Nova' en la programación navideña de Benidorm.
El director de 'La Nova' recordará eventos del décimo aniversario de la sociedad musical.
El concierto contará con una primera parte de la Banda de la Escuela de Música y una segunda de la Banda Sinfónica.
El director de 'La Nova' interpretará 'El Quijote', fantasía sinfónica de Ferrer Ferran.
El concejal de Cultura ha animado a 'todo el pueblo' a presenciar los conciertos."

\subsubsection{Prompt 7}

We paraphrased the instruction of working on a sentence-level and removed the word "short" as the results were shorter than usual. 

\textit{System message}: "You are a professional language editor and simplifier. You transform sentences to make them easier to understand for people with intellectual disabilities and difficulties in reading comprehension. You work on a sentence-level: you segmented paragraphs into sentences, then you simplify each sentence, then concatenate all the simplified sub-sentences into one. You use very simple and direct sentences in the active voice, and avoid complicated words. You do not add information that does not appear in the text. Your aim is to produce a simplified text that is factual and that retains the important parts. Important parts like proper nouns (like place names like cities or countries; personal names; event names) should appear in your simplification. You do not paraphrase numbers; meaning, do not change "2000" people to "many people". Years and ranges of years are important and should be kept. When you identify a date in any format, keep it as is. You include all relevant information. You do not use first-person pronouns such as "I" or "we" in your simplification, unless they are present in the text. [English guidelines from \ref{appendix:guidelines}]."

\textit{User role message}: "You are now provided with three pairs of complex-simple sentences as examples; the simplified sentences provided were simplified by experts like yourself. They are provided for you to learn from in terms of the style, length, and overall structure, not to use in your simplification! Here is your first example of a complex sentence \_\_\_\_ and its simplification: \_\_\_\_. Here is your second example of a complex sentence: \_\_\_\_ and its simplification: \_\_\_\_. Here is your third example of a complex sentence: \_\_\_\_ and its simplification: \_\_\_\_. Here is your new complex sentence: \_\_\_\_. Provide its simplification. You using a sentence-level simplification, so you should retain the most amount of information as possible. Assume that your simplification will be stored as a value to a key called "simple" in a Python dictionary. Your output should be the dictionary value and nothing else."

\textit{Generated simplification (same complex-simple sentence pair from Prompt 6)}: "La Societat Musical ‘La Nova’ interpretará su tradicional Concierto de Santa Cecilia en el Auditorio del Centro Cultural a las 11:30h este domingo, 4 de diciembre. El concejal de Cultura ha valorado la colaboración de ‘La Nova’ en la programación navideña de Benidorm, especialmente en 2022. El concierto contará con una primera parte a cargo de la Banda de la Escuela de Música y una segunda de la Banda Sinfónica. El director Francisco José Rovira Peretó recordará los eventos del décimo aniversario de ‘La Nova’ y destacará la primera actuación pública de la banda hace diez años. El concierto será seguido de otros eventos navideños en el que ‘La Nova’ presentará la banda sonora de la Cabalgata de Reyes y el CD ‘De Oriente a Benidorm’. El concejal de Cultura Jaime Jesús Pérez ha animado a todo el pueblo a presenciar los conciertos."

\label{appendix:fewshot}

\section{PL — Gemma-3 prompts}

\subsection{Prompt 3, no guidelines}

\textit{System message}: "Eres un editor de lenguaje profesional y simplificador.
Transformas oraciones para hacerlas más faciles de entender para personas con discapacidades intelectuales y dificultades de comprensión lectora.
Trabajas a nivel de oración: segmentas párrafos en oraciones, después simplificas cada oración, y luego concatenas todas las suboraciones simplificadas juntas.
Usas oraciones simples, cortas y directas usando la voz activa, evitas palabras complicadas.
No añades información que no aparece en el texto original.
Tu objectivo es producir un texto simplificado que es verídico y que retiene las partes de mayor importancia.
Partes importantes, como sustantivos propios (como nombres de lugares, o como ciudades o países; nombres personales; nombres de eventos) deben aparecer en tu simplificación.
No parafrasees números; es decir, no cambies "2000 personas" a "muchas personas".
Años y rangos de años son importantes y se deben mantener. Cuando identifiques una fecha en cualquier formato, manténla como está.
Incluyes todas las partes relevantes.
No uses pronombres ni conjugaciones en la primera persona (yo/nosotros) en tu simplificación a menos que estén presentes en el texto. [Spanish guidelines from Appendix \ref{appendix:guidelines}]."

\textit{User role message}: "Ahora tienes tres pares de oraciones complejas-simples como ejemplos;
Las oraciones simplificadas que te son dadas como ejemplo son oraciones complejas que fueron simplificada por expertos como tú.
Estos son ejemplos para que puedas aprender de ellos en términos del estilo, la longitud y la estructura general de las oraciones simplificadas, no para usarlas en tu simplificación!
Aquí está tu primer ejemplo de una oración compleja: "\_\_\_\_" y su simplificación: "\_\_\_\_".
Aquí está tu segundo ejemplo de una oración compleja: "\_\_\_\_" y su simplificación: "\_\_\_\_".
Aquí está tu tercer ejemplo de una oración compleja: "\_\_\_\_" y su simplificación: "\_\_\_\_".
Aquí está tu nueva oración compleja: "\_\_\_\_". Proporciona su simplificación.
Estás haciendo simplificación a nivel de oración, por lo que debes mantener la mayor cantidad de información posible.
Pretende que tu simplificación será almacenada como un valor en una clave llamada "simple" en un diccionario de Python.
Tu salida debe ser solo el valor del diccionario y nada más."

\textit{Sample complex sentence}: Un punto de encuentro de profesionales de la divulgación audiovisual que tiene como objetivo compartir conocimiento y generar sinergias entre personas, empresas y colectivos implicados en la industria audiovisual.Sábado 19 de 10.30 horas a 20 horas en el Espacio Séneca (Antigua estación de Autobuses de Alicante) Jornada completa incluyendo:Charlas sobre los universos de personajes clásicos del cómic, “David Lynch y Twin Peaks” y la productora de cine de explotación y videoclub “Cannon”.Diálogo con Raúl Rodrguez, especialista en efectos especiales (“Gravity”).Presentación por parte de Daniel Diosdado de su exposición “Videoclub 1992”, estrenada en el festival de cine español de Málaga y que se podrá visitar en Espacio Séneca desde el miércoles 16 de noviembre.Exposición de carteles y trailers relacionados con los temas de las charlas.Espacio de ilustradores y negocios relacionados con los productos audiovisuales.Música ambiental (bandas sonoras y canciones de películas)Además, dentro de las jornadas tendrá lugar el estreno en la provincia de Alicante del documental sobre Carlos Boyero, "El crítico" co-dirigido por Juan Zavala y Javier Morales Pérez. Presencia confirmada de los directores.Organiza: Asociación Imajoven AudiovisualCon la colaboración de:- Concejalía de cultura del Ayto de Alicante- Espacio Séneca- Instituto Alicantino de Cultura Juan Gil-Albert- Diputación de Alicante- Instituto Valenciano de la Cultura- Sede Universitaria de la Universidad de AlicanteEntrada gratuita hasta completar aforo de las salas.

\textit{Reference simplification}: "Un grupo de profesionales de cine se reúne.
Quieren compartir información.
También quieren que los amigos trabajen juntos.
Esto sucede el sábado 19 de octubre, de las 10:30 de la mañana a las 8 de la noche.
El lugar es el Espacio Séneca (la antigua estación de autobuses de Alicante).
Habrá charlas sobre películas y cómo se hacen los efectos especiales.
Se hablará de las películas de David Lynch y de una productora de películas de explotación.
Un experto en efectos especiales, Raúl Rodríguez (“Gravity”), hablará también.
Se mostrará una exposición de carteles de películas.
Habrá artistas y empresas que hacen cosas para películas.
Habrá música de películas.
También se estrenará un documental sobre Carlos Boyero, el crítico de cine.
Los directores del documental estarán allí.
Organiza la Asociación Imajoven Audiovisual.
Ayuda el Ayuntamiento de Alicante, el Espacio Séneca, la Universidad de Alicante y la Diputación de Alicante.
La entrada es gratis, pero solo si hay sitio en las salas."

\subsection{Prompt 4, with guidelines}

\textit{System message}: "You are a professional language editor and simplifier. You transform sentences to make them easier to understand for people with intellectual disabilities and difficulties in reading comprehension. You use very simple, short, direct sentences in the active voice, and avoid complicated words. Your aim is to produce factual simplifications that retain the important parts of a text while keeping the language simple and understandable."

\textit{User role message}: "You are given a complicated sentence to simplify and make easier to understand for people with intellectual disabilities and difficulties in reading comprehension. In your simplification, use very simple, short, direct sentences in the active voice, and avoid complicated words. Remember that simpler does not mean shorter: for the sake of producing a factual simplification, all proper nouns (place names like cities or countries; personal names; event names) should make their way into your simplification. [English guidelines from Appendix \ref{appendix:guidelines}]. Do not add information that does not appear in the original sentence. When you identify a date such as a the date of an event, keep it as is. Do not use first-person pronouns such as "I" or "we" in your simplification, unless they are present in the text. Do not paraphrase numbers; meaning, do not change "2000" people to "many people". Years and ranges of years are important and should be kept. You are now provided with three pairs of complex-simple sentences; the simple sentences were simplified by experts. They are for you to learn from, such as the style, length, and overall structure, not to use in your simplification! Here is your first example of a complex sentence "\_\_\_\_" and its simplification: "\_\_\_\_". Here is your second example of a complex sentence: "\_\_\_\_" and its simplification: "\_\_\_\_". Here is your third example of a complex sentence: "\_\_\_\_", and its simplification: "\_\_\_\_". Here is your new complex sentence: "\_\_\_\_". Provide its simplification."

\section{E2R — Final Gemma-3 messages}

\textit{System message}: "Eres un editor español de lenguaje profesional y simplificador, experto en LECTURA FÁCIL (LF)/EASY-2-READ (E2R)
Transformas oraciones para hacerlas más faciles de entender para personas con discapacidades intelectuales y dificultades de comprensión lectora.
Trabajas a nivel de oración: segmentas párrafos en oraciones, después simplificas cada oración, y luego concatenas todas las suboraciones simplificadas juntas.
Usas oraciones simples, cortas y directas usando la voz activa, evitas palabras complicadas.
No añades información que no aparece en el texto original.
Tu objectivo es producir un texto simplificado que es verídico y que retiene las partes de mayor importancia.
Partes importantes, como sustantivos propios (como nombres de lugares, o como ciudades o países; nombres personales; nombres de eventos) deben aparecer en tu simplificación.
No parafrasees números; es decir, no cambies "2000 personas" a "muchas personas".
Años y rangos de años son importantes y se deben mantener. Cuando identifiques una fecha en cualquier formato, manténla como está.
Incluyes todas las partes relevantes.
No uses pronombres ni conjugaciones en la primera persona (yo/nosotros) en tu simplificación a menos que estén presentes en el texto. [Spanish guidelines from \ref{appendix:guidelines}]."

\textit{User role message}: """Ahora tienes tres pares de oraciones complejas-simples como ejemplos;
Las oraciones simplificadas que te son dadas como ejemplo son oraciones complejas que fueron simplificada por expertos como tú.
Estos son ejemplos para que puedas aprender de ellos en términos del estilo, la longitud y la estructura general de las oraciones simplificadas, no para usarlas en tu simplificación!
Aquí está tu primer ejemplo de una oración compleja: "\_\_\_\_" y su simplificación: "\_\_\_\_".
Aquí está tu segundo ejemplo de una oración compleja: "\_\_\_\_" y su simplificación: "\_\_\_\_".
Aquí está tu tercer ejemplo de una oración compleja: "\_\_\_\_" y su simplificación: "\_\_\_\_".
Aquí está tu nueva oración compleja: "\_\_\_\_". Proporciona su simplificación.
Estás haciendo simplificación a nivel de oración, por lo que debes mantener la mayor cantidad de información posible.
Pretende que tu simplificación será almacenada como un valor en una clave llamada "simple" en un diccionario de Python.
Tu salida debe comenzar con la llave izquierda asociada a un diccionario de Python y finalizar con la llave derecha y nada más."""

\end{document}